\documentclass[sigconf]{acmart}

\usepackage{makecell}
\usepackage{booktabs,tabularx}
\usepackage{graphicx}
\usepackage{tikz}
\usepackage{placeins}
\PassOptionsToPackage{protrusion=true,expansion=true,factor=1200,stretch=40,shrink=40}{microtype}
\usetikzlibrary{arrows.meta,positioning}
\newcolumntype{L}[1]{>{\raggedright\arraybackslash}p{#1}}
\newcolumntype{Y}{>{\raggedright\arraybackslash}X}

\newcommand{\probmarker}{\tikz[baseline=-0.45ex,x=0.12em,y=0.12em]{\draw[line width=0.35pt] (0,1) -- (1,0) -- (0,-1) -- (-1,0) -- cycle;}}

\setcopyright{none}
\renewcommand\footnotetextcopyrightpermission[1]{}

\acmConference[ACM AI Summit '26]{ACM AI Leadership Summit 2026}{August 30--September 2, 2026}{Atlanta, GA, USA}
\acmYear{2026}
\acmBooktitle{Proceedings of the ACM AI Leadership Summit 2026 (ACM AI Summit '26), August 30--September 2, 2026, Atlanta, GA, USA}

\title[Tracing the Evidence Behind Zero-Shot TSF]{Tracing the Evidence Behind Zero-Shot Time-Series Forecasting: A Source-First Taxonomy and Audit Framework}

\author{Delun Kong}
\email{delun.kong@tum.de}
\affiliation{%
  \institution{Department of Operations and Technology\\Technical University of Munich}
  \city{Heilbronn}
  \country{Germany}}

\author{Wanyun Ling}
\email{wanyun.ling@tum.de}
\affiliation{%
  \institution{Department of Operations and Technology\\Technical University of Munich}
  \city{Heilbronn}
  \country{Germany}}

\author{Chenxi Liu}
\email{chenxi.liu@cair-cas.org.hk}
\affiliation{%
  \institution{Hong Kong Institute of Science \& Innovation\\Chinese Academy of Sciences}
  \city{Hong Kong}
  \country{China}
  }

\author{Ziyue Li}
\email{ziyue.li@tum.de}
\affiliation{%
  \institution{Department of Operations and Technology, Heilbronn Data Science Center, Munich Data Science Institute\\Technical University of Munich}
  \city{Heilbronn}
  \country{Germany}}

\begin{document}

\begin{abstract}
Zero-shot time-series forecasting (TSF) is often described as forecasting without target-specific parameter updates, but that training-status condition does not specify what evidence the system may use. A frozen language model prompted with serialized values, a time-series model pretrained on broad forecasting corpora, and a retrieval-augmented forecaster may all satisfy the no-update condition while drawing on different transferable evidence. This paper argues that zero-shot TSF should therefore be governed as an evidence-access claim. We propose a source-first taxonomy that separates three primary evidence sources---frozen LLM prior reuse, parametric time-series pretraining, and retrieval-augmented external memory---from the architectures that implement them. After the source is identified, four additional audit questions remain: task interface, forecast object and scoring, prediction-time context, and resource budget. The resulting agenda is to make zero-shot leaderboards auditable by reporting evidence boundaries and interface assumptions alongside scores, so that benchmark progress reflects transferable forecasting capability rather than undisclosed changes in context, memory, or budget.
\end{abstract}



\maketitle

\begin{figure*}[t]
\centering
\resizebox{0.98\textwidth}{!}{%
\begingroup
\pgfdeclarelayer{taxonomybg}
\pgfsetlayers{taxonomybg,main}
\definecolor{FrozenMain}{HTML}{6F82AD}
\definecolor{FrozenFill}{HTML}{F4F6FB}
\definecolor{ParamMain}{HTML}{789678}
\definecolor{ParamFill}{HTML}{F3F7F2}
\definecolor{RetrMain}{HTML}{B9825B}
\definecolor{RetrFill}{HTML}{FBF4EF}
\definecolor{NeutralFill}{HTML}{F6F6F3}
\definecolor{NeutralBorder}{HTML}{A6A7A2}
\definecolor{MechanismFill}{HTML}{FEFEFC}
\definecolor{PaperFill}{HTML}{FFFFFF}
\definecolor{TaxConnector}{HTML}{B6B7B2}
\newcommand{\sourcecompact}[3]{{\bfseries\color{#1}\strut #2}\\[-1pt]{\fontsize{6.2pt}{6.8pt}\selectfont\itshape\color{black!66}\strut #3}}
\newcommand{\ragtitle}{\resizebox{3.00cm}{!}{\bfseries Retrieval-Augmented Memory}}
\newcommand{\oneline}[1]{\mbox{#1}}
\begin{tikzpicture}[
  x=1cm,
  y=1cm,
  font=\rmfamily,
  connector/.style={draw=TaxConnector, line width=0.40pt, rounded corners=2pt},
  rootnode/.style={draw=NeutralBorder, line width=0.45pt, rounded corners=2.5pt,
    fill=NeutralFill, align=center, text width=1.45cm, minimum height=0.90cm,
    inner xsep=4pt, inner ysep=3pt, font=\bfseries\fontsize{7.7pt}{8.4pt}\selectfont,
    text=black!88},
  sourcenode/.style={line width=0.50pt, rounded corners=2.5pt, align=center,
    text width=3.40cm, minimum width=3.58cm, minimum height=0.72cm,
    inner xsep=5pt, inner ysep=2pt,
    font=\fontsize{7.1pt}{7.7pt}\selectfont, text=black!88},
  sourceFrozen/.style={sourcenode, draw=FrozenMain, fill=FrozenFill},
  sourceParametric/.style={sourcenode, draw=ParamMain, fill=ParamFill},
  sourceRetrieval/.style={sourcenode, draw=RetrMain, fill=RetrFill},
  mechanism/.style={line width=0.38pt, rounded corners=2.3pt, align=center,
    text width=3.70cm, minimum width=3.86cm, minimum height=0.42cm,
    inner xsep=4pt, inner ysep=1.6pt,
    fill=MechanismFill, font=\fontsize{6.6pt}{7.2pt}\selectfont, text=black!82},
  mechanismFrozen/.style={mechanism, draw=FrozenMain!45},
  mechanismParametric/.style={mechanism, draw=ParamMain!45},
  mechanismRetrieval/.style={mechanism, draw=RetrMain!45},
  papernode/.style={line width=0.34pt, rounded corners=2.1pt, align=center,
    text width=4.92cm, minimum width=5.08cm, minimum height=0.42cm,
    inner xsep=4pt, inner ysep=1.6pt,
    fill=PaperFill, font=\fontsize{6.3pt}{6.9pt}\selectfont, text=black!78},
  paperFrozen/.style={papernode, draw=FrozenMain!35},
  paperParametric/.style={papernode, draw=ParamMain!35},
  paperRetrieval/.style={papernode, draw=RetrMain!35}
]

\node[rootnode] (root) at (0.70,-2.64) {Zero-Shot\\Time\\Series\\Forecasting};

\node[sourceFrozen] (llm) at (4.10,-0.68)
  {\sourcecompact{FrozenMain}{\oneline{Frozen LLM Prior Reuse}}{\oneline{frozen priors; no TS-specific training}}};
\node[sourceParametric] (param) at (4.10,-2.64)
  {\sourcecompact{ParamMain}{\oneline{Parametric Pretraining}}{\oneline{zero-shot ability in model parameters}}};
\node[sourceRetrieval] (rag) at (4.10,-4.60)
  {\sourcecompact{RetrMain}{\ragtitle}{\oneline{adaptation via retrieved patterns}}};

\node[mechanismFrozen] (ser) at (8.70,-0.42) {\oneline{Numeric Serialization}};
\node[mechanismFrozen] (prompt) at (8.70,-0.94) {\oneline{Forecasting-Aware Prompting}};

\node[mechanismParametric] (numpre) at (8.70,-2.12) {\oneline{Numerical Time-Series Pretraining}};
\node[mechanismParametric] (bias) at (8.70,-2.64) {\oneline{Inductive-Bias-Oriented Pretraining}};
\node[mechanismParametric] (crossmodal) at (8.70,-3.16) {\oneline{Cross-Modal Alignment Pretraining}};

\node[mechanismRetrieval] (learn) at (8.70,-4.08) {\oneline{Learnable Retrieval}};
\node[mechanismRetrieval] (fusion) at (8.70,-4.60) {\oneline{Retrieved Pattern Fusion}};
\node[mechanismRetrieval] (interact) at (8.70,-5.12) {\oneline{Query-Guided Retrieval Filtering}};

\node[paperFrozen] (serp) at (13.98,-0.42) {\oneline{\itshape LLMTime}};
\node[paperFrozen] (promptp) at (13.98,-0.94) {\oneline{\itshape LSTPrompt}};

\node[paperParametric] (numprep) at (13.98,-2.12)
  {\oneline{\itshape Lag-Llama, TimesFM, Chronos, MOIRAI, Time-MoE}};
\node[paperParametric] (biasp) at (13.98,-2.64)
  {\oneline{\itshape TTM, Mamba4Cast, TiRex, Reverso}};
\node[paperParametric] (crossmodalp) at (13.98,-3.16) {\oneline{\itshape ChatTime}};

\node[paperRetrieval] (learnp) at (13.98,-4.08) {\oneline{\itshape TimeRAF}};
\node[paperRetrieval] (fusionp) at (13.98,-4.60) {\oneline{\itshape TS-RAG}};
\node[paperRetrieval] (interactp) at (13.98,-5.12) {\oneline{\itshape Cross-RAG}};

\begin{pgfonlayer}{taxonomybg}
\coordinate (rootJoin) at (2.05,-2.64);
\coordinate (sourceSpineTop) at (2.05,-0.68);
\coordinate (sourceSpineBottom) at (2.05,-4.60);
\draw[connector] (root.east) -- (rootJoin);
\draw[connector] (sourceSpineTop) -- (sourceSpineBottom);
\draw[connector] (sourceSpineTop) -- (llm.west);
\draw[connector] (rootJoin) -- (param.west);
\draw[connector] (sourceSpineBottom) -- (rag.west);

\coordinate (llmJoin) at (6.14,-0.68);
\coordinate (llmSpineTop) at (6.14,-0.42);
\coordinate (llmSpineBottom) at (6.14,-0.94);
\draw[connector] (llm.east) -- (llmJoin);
\draw[connector] (llmSpineTop) -- (llmSpineBottom);
\draw[connector] (llmSpineTop) -- (ser.west);
\draw[connector] (llmSpineBottom) -- (prompt.west);

\coordinate (paramJoin) at (6.14,-2.64);
\coordinate (paramSpineTop) at (6.14,-2.12);
\coordinate (paramSpineBottom) at (6.14,-3.16);
\draw[connector] (param.east) -- (paramJoin);
\draw[connector] (paramSpineTop) -- (paramSpineBottom);
\draw[connector] (paramSpineTop) -- (numpre.west);
\draw[connector] (paramJoin) -- (bias.west);
\draw[connector] (paramSpineBottom) -- (crossmodal.west);

\coordinate (ragJoin) at (6.14,-4.60);
\coordinate (ragSpineTop) at (6.14,-4.08);
\coordinate (ragSpineBottom) at (6.14,-5.12);
\draw[connector] (rag.east) -- (ragJoin);
\draw[connector] (ragSpineTop) -- (ragSpineBottom);
\draw[connector] (ragSpineTop) -- (learn.west);
\draw[connector] (ragJoin) -- (fusion.west);
\draw[connector] (ragSpineBottom) -- (interact.west);

\coordinate (paperSpineX) at (10.94,0);
\foreach \m/\p/\y in {ser/serp/-0.42,prompt/promptp/-0.94,numpre/numprep/-2.12,bias/biasp/-2.64,crossmodal/crossmodalp/-3.16,learn/learnp/-4.08,fusion/fusionp/-4.60,interact/interactp/-5.12} {
  \draw[connector] (\m.east) -- (10.94,\y) -- (\p.west);
}

\end{pgfonlayer}

\end{tikzpicture}
\endgroup%
}
\caption{Source-first audit taxonomy for zero-shot TSF. Branches identify primary evidence sources, assess mechanisms, and representative methods; hybrid methods are placed by the primary evidence source active in the reported configuration.}
\Description{A tree diagram classifying zero-shot time-series forecasting by source of generalization, mechanism, and representative methods.}
\label{fig:taxonomy}
\end{figure*}
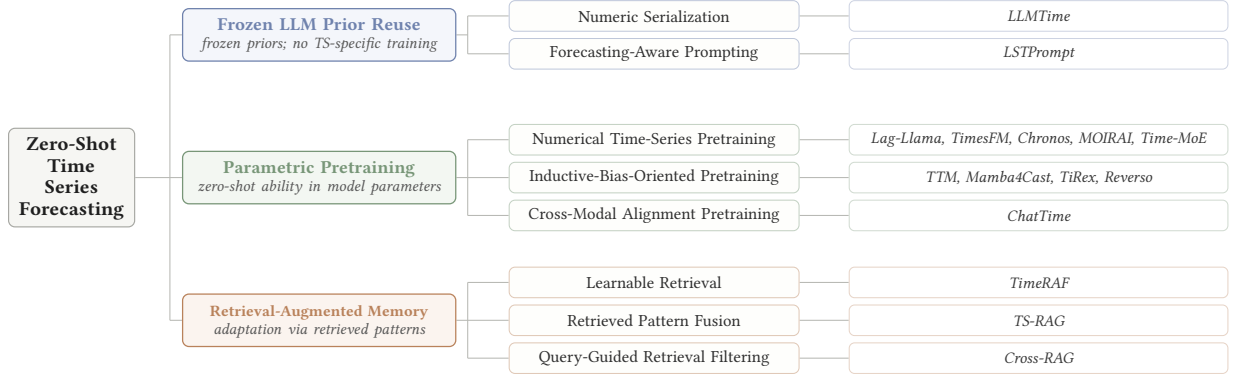

\section{Introduction}

In time-series forecasting (TSF), the zero-shot label usually specifies that model parameters are not updated on the target series. It does not specify the evidence boundary: the information available before evaluation through pretrained parameters or during prediction through prompts, covariates, or retrieval. This omission matters in a foundation-model setting \cite{bommasani2021foundation}. LLMTime exposes a frozen language-model prior through numerical serialization \cite{gruver2024llmtime}; Chronos stores transferable forecasting structure in parameters learned from broad time-series corpora \cite{ansari2024chronos}; and TimeRAF changes prediction-time evidence by retrieving from external time-series memory \cite{zhang2025timeraf}. These systems can all be reported as zero-shot, but their scores answer different evidence-access questions.

Prior surveys organize time-series foundation models around methodological components, data category, model families, scope, modality and downstream task \cite{liang2024foundation,jin2026large}. Unlike architecture-centred surveys, our taxonomy separates the implementation mechanism from the evidence available to each reported forecasting configuration. Architecture still matters: mixers \cite{ekambaram2024ttm}, state-space models \cite{bhethanabhotla2024mamba4cast}, xLSTM blocks \cite{auer2025tirex}, and hybrid convolution--RNN layers \cite{fu2026reverso} all shape transfer behavior. But architecture alone does not determine whether a score reflects frozen LLM reuse, parametric time-series pretraining, or retrieval-augmented memory. Fig.~\ref{fig:taxonomy} therefore starts from the primary evidence source and places mechanisms and representative methods beneath it.

Even after the primary evidence source is fixed, comparisons remain fragile: point-error metrics are not interchangeable across settings \cite{hyndman2006accuracy}, probabilistic forecasts require scoring rules matched to the forecast object \cite{gneiting2007proper}, and reported zero-shot scores can answer different questions.

This paper makes three contributions. First, it proposes a source-first taxonomy that classifies zero-shot TSF systems by the primary source of transferable evidence rather than by architecture alone. Second, it identifies four cross-cutting audit conditions---task interface, forecast object and scoring, prediction-time context, and resource budget---that remain after the evidence source is identified. Third, it turns these conditions into a minimum disclosure checklist and a benchmark governance agenda for evidence-boundary reporting.

\section{Sources of Zero-Shot Generalization}

\begin{figure*}[t]
\centering
\resizebox{0.98\textwidth}{!}{%
\begingroup
\definecolor{branchBlue}{RGB}{62,105,169}
\definecolor{branchBlueFill}{RGB}{245,248,253}
\definecolor{branchBlueBand}{RGB}{250,252,255}
\definecolor{branchGreen}{RGB}{70,132,83}
\definecolor{branchGreenFill}{RGB}{246,251,247}
\definecolor{branchGreenBand}{RGB}{250,253,250}
\definecolor{branchOrange}{RGB}{198,111,54}
\definecolor{branchOrangeFill}{RGB}{255,249,245}
\definecolor{branchOrangeBand}{RGB}{255,252,249}
\definecolor{axisGray}{RGB}{160,160,160}
\definecolor{textInk}{RGB}{35,38,42}
\newcommand{\modelentry}[2]{\strut{\bfseries #1}\\[-1pt]{\fontsize{5.2pt}{5.8pt}\selectfont\strut\mbox{#2}}}
\begin{tikzpicture}[x=1cm,y=1cm,font=\rmfamily]
\tikzset{
  lane/.style={rounded corners=3pt, line width=0.25pt, draw=black!6},
  laneLabel/.style={font=\bfseries\scriptsize, align=right, text width=2.10cm, inner sep=1pt, anchor=east},
  laneLabelLong/.style={font=\bfseries\fontsize{5.0pt}{5.7pt}\selectfont,
    align=center, text width=1.92cm, inner sep=1pt, anchor=center},
  modelcard/.style={rounded corners=2.5pt, line width=0.55pt, align=center,
    text width=1.92cm, minimum width=2.04cm, minimum height=0.50cm,
    inner xsep=3pt, inner ysep=1.5pt, font=\scriptsize, text=textInk},
  bluecard/.style={modelcard, draw=branchBlue, fill=branchBlueFill},
  greencard/.style={modelcard, draw=branchGreen, fill=branchGreenFill},
  orangecard/.style={modelcard, draw=branchOrange, fill=branchOrangeFill},
  marker/.style={font=\scriptsize, inner sep=0pt, anchor=north east, xshift=-1.5pt, yshift=-1.0pt},
  yearTick/.style={draw=axisGray, line width=0.35pt},
  yearLabel/.style={font=\scriptsize, text=black!70},
  legendText/.style={font=\scriptsize, text=black!72, inner sep=0pt}
}

\coordinate (x2023) at (3.45,0);
\coordinate (x2024) at (6.80,0);
\coordinate (x2025) at (11.10,0);
\coordinate (x2026) at (14.75,0);

\node[font=\bfseries\large, text=textInk] at (8.45,2.92)
  {Evolution Timeline of Zero-Shot Time-Series Forecasting};
\node[font=\itshape\scriptsize, text=branchBlue] at (3.70,2.55) {Reused Priors};
\draw[-{Latex[length=1.6mm]}, axisGray, line width=0.35pt] (4.55,2.55) -- (6.28,2.55);
\node[font=\itshape\scriptsize, text=branchGreen] at (7.45,2.55) {Pretrained TSFMs};
\draw[-{Latex[length=1.6mm]}, axisGray, line width=0.35pt] (8.58,2.55) -- (10.45,2.55);
\node[font=\itshape\scriptsize, text=branchOrange] at (12.30,2.55) {Retrieval-Augmented Memory};

\fill[lane, fill=branchBlueBand] (0.55,1.64) rectangle (15.95,2.20);
\fill[lane, fill=branchGreenBand] (0.55,0.08) rectangle (15.95,1.32);
\fill[lane, fill=branchOrangeBand] (0.55,-0.65) rectangle (15.95,-0.10);

\node[laneLabel, text=branchBlue] at (2.25,1.92) {Frozen LLM\\Prior Reuse};
\node[laneLabel, text=branchGreen] at (2.25,0.70) {Parametric\\Pretraining};
\node[laneLabelLong, text=branchOrange] at (1.70,-0.38) {Retrieval-Augmented\\Memory};

\node[bluecard] (llmtime) at (3.45,1.92)
  {\modelentry{LLMTime}{digital strings}};
\node[marker, text=branchBlue] at (llmtime.north east) {\probmarker};
\node[bluecard] (lstprompt) at (6.80,1.92)
  {\modelentry{LSTPrompt}{long-short prompts}};
\node[marker, text=branchBlue] at (lstprompt.north east) {$\square$};

\node[greencard] (lagllama) at (3.45,1.02)
  {\modelentry{Lag-Llama}{lag covariates}};
\node[marker, text=branchGreen] at (lagllama.north east) {\probmarker};
\node[greencard] (timesfm) at (4.75,0.40)
  {\modelentry{TimesFM}{input/output patches}};
\node[marker, text=branchGreen] at (timesfm.north east) {$\square$};

\node[greencard] (ttm) at (5.90,1.02)
  {\modelentry{TTM}{adaptive patching}};
\node[marker, text=branchGreen] at (ttm.north east) {$\square$};
\node[greencard] (chronos) at (7.10,0.40)
  {\modelentry{Chronos}{quantized tokens}};
\node[marker, text=branchGreen] at (chronos.north east) {\probmarker};
\node[greencard] (moirai) at (8.25,1.02)
  {\modelentry{MOIRAI}{masked patches}};
\node[marker, text=branchGreen] at (moirai.north east) {\probmarker};
\node[greencard] (chattime) at (9.65,0.40)
  {\modelentry{ChatTime}{marked value tokens}};
\node[marker, text=branchGreen] at (chattime.north east) {$\square$};
\node[greencard] (timemoe) at (10.95,1.02)
  {\modelentry{TIME-MOE}{sparse MoE}};
\node[marker, text=branchGreen] at (timemoe.north east) {$\square$};
\node[greencard] (mamba) at (12.25,0.40)
  {\modelentry{Mamba4Cast}{SSM}};
\node[marker, text=branchGreen] at (mamba.north east) {$\square$};

\node[greencard] (tirex) at (13.35,1.02)
  {\modelentry{TiRex}{xLSTM + CPM}};
\node[marker, text=branchGreen] at (tirex.north east) {\probmarker};
\node[greencard] (reverso) at (14.85,0.40)
  {\modelentry{Reverso}{long conv + DeltaNet}};
\node[marker, text=branchGreen] at (reverso.north east) {$\square$};

\node[orangecard] (timeraf) at (6.80,-0.38)
  {\modelentry{TimeRAF}{channel prompts}};
\node[marker, text=branchOrange] at (timeraf.north east) {$\square$};
\node[orangecard] (tsrag) at (11.10,-0.38)
  {\modelentry{TS-RAG}{retrieved pattern fusion}};
\node[marker, text=branchOrange] at (tsrag.north east) {$\circ$};
\node[orangecard] (crossrag) at (14.75,-0.38)
  {\modelentry{Cross-RAG}{query--retrieval cross-attn}};
\node[marker, text=branchOrange] at (crossrag.north east) {$\circ$};

\draw[-{Latex[length=1.7mm]}, axisGray, line width=0.45pt] (3.15,-0.90) -- (15.65,-0.90);
\foreach \x/\yr in {3.45/2023,6.80/2024,11.10/2025,14.75/2026} {
  \draw[yearTick] (\x,-0.83) -- (\x,-0.97);
  \node[yearLabel] at (\x,-1.17) {\yr};
}

\node[draw=black!10, fill=black!2, rounded corners=2pt,
  minimum width=14.0cm, minimum height=0.58cm] at (8.45,-1.82) {};
\node[legendText, anchor=west] at (1.55,-1.82) {\textbf{Color:}};
\draw[draw=branchBlue, fill=branchBlueFill, line width=0.45pt] (2.38,-1.89) rectangle (2.54,-1.73);
\node[legendText, anchor=west] at (2.62,-1.82) {Frozen LLM};
\draw[draw=branchGreen, fill=branchGreenFill, line width=0.45pt] (4.24,-1.89) rectangle (4.40,-1.73);
\node[legendText, anchor=west] at (4.48,-1.82) {Parametric};
\draw[draw=branchOrange, fill=branchOrangeFill, line width=0.45pt] (5.98,-1.89) rectangle (6.14,-1.73);
\node[legendText, anchor=west] at (6.22,-1.82) {Retrieval};
\node[legendText, anchor=west] at (7.78,-1.82) {\textbf{Output:} \probmarker{} Prob. \quad $\square$ Point \quad $\circ$ Mixed};
\node[legendText, anchor=west] at (11.48,-1.82) {\textbf{Micro-label:} interface/mechanism};

\end{tikzpicture}
\endgroup%
}
\caption{Chronology of representative zero-shot TSF methods by evidence source. Rows and colors follow the source-first branches in Fig.~\ref{fig:taxonomy}. The timeline highlights the widening evidence-access surface from reused LLM priors to pretrained TSFMs and retrieval-augmented memory.}
\Description{A chronological timeline of representative zero-shot time-series forecasting methods, grouped by taxonomy branch and output type.}
\label{fig:timeline}
\end{figure*}
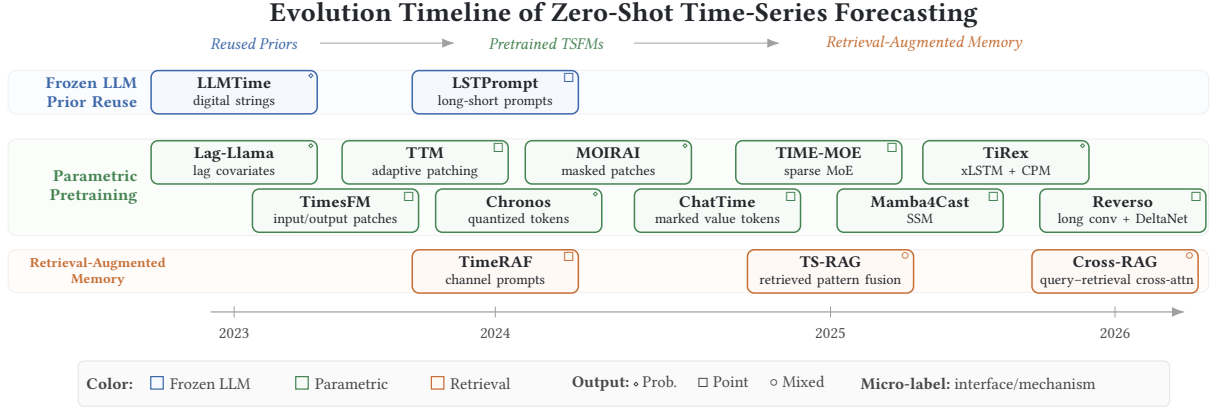

Fig.~\ref{fig:taxonomy} organizes methods by the primary evidence source behind a zero-shot claim rather than by model architecture. We consider reported zero-shot forecasting configurations whose training and inference procedures provide sufficient information to identify the primary evidence source. We use \emph{evidence boundary} to denote the information sources available to a forecasting system, either before evaluation through pretrained parameters or during evaluation through prompting, covariates, or retrieval. The top level of the taxonomy separates where transferable evidence is stored or accessed: in a frozen language-model prior, in parameters learned through time-series pretraining, or in an external memory retrieved at inference time. The second level asks how that evidence enters the forecast. Numeric serialization and forecasting-aware prompts expose a frozen LLM prior; broad real or synthetic time-series pretraining stores forecasting structure in model weights; and retrieval-augmented methods add external examples or patterns at prediction time. Classification is applied to the reported forecasting configuration rather than to the model family. A configuration is retrieval-augmented when retrieved external examples materially condition its prediction. Without retrieval, time-series forecasting pretraining places it in the parametric branch; an unchanged general-purpose LLM accessed only through serialization or prompting belongs to frozen-prior reuse. Additional active channels are reported as secondary evidence. Fig.~\ref{fig:timeline} places representative methods chronologically under this source-first view.

\section{Frozen LLM Prior Reuse}

The frozen-LLM branch asks whether a model trained primarily on text can reuse its language-model prior for numerical sequence continuation without target-specific updating.

\subsection{Numeric Serialization}

Numeric serialization puts numerical histories into textual form and decodes textual continuations back into numerical forecasts. LLMTime is the clearest example: it encodes values as digit strings, treats forecasting as next-token continuation, and converts token-level distributions into numerical samples, point summaries, or likelihoods \cite{gruver2024llmtime}. It belongs in the frozen-LLM branch because the transferable evidence is the pretrained language-model prior; serialization is the task interface that makes this prior usable for forecasting.

\subsection{Forecasting-Aware Prompting}

Forecasting-aware prompting changes the interface rather than relying on raw numerical continuation alone. LSTPrompt decomposes forecasting into short-term and long-term subtasks and designs prompts for each, aiming to improve zero-shot adaptation without changing the underlying pretrained model \cite{liu2024lstprompt}. Its source remains the frozen language-model prior because prompting changes the task interface, not the parameter store or the evidence source.

\section{Parametric Pretraining}

Parametric pretraining treats zero-shot TSF as the reuse of forecasting-relevant structure already encoded in model weights through broad time-series, synthetic, or time-series--text pretraining.

\subsection{Numerical Time-Series Pretraining}

Numerical time-series pretraining learns forecasting behavior from broad real and/or synthetic time-series distributions. Lag-Llama, TimesFM, Chronos, MOIRAI, and Time-MoE fit this mechanism because their zero-shot forecasts rely on structure learned before target evaluation from large pretraining corpora \cite{rasul2024lagllama,das2024timesfm,ansari2024chronos,woo2024moirai,shi2025timemoe}. Their differences in lag features, patching, quantization, unified training, and sparse-expert scaling are important model-specific mechanisms, but they do not change the primary evidence source: transferable forecasting evidence stored in model parameters. For example, Chronos is classified as parametric pretraining because its transferable forecasting evidence is encoded in parameters learned from broad time-series corpora; quantization is an interface mechanism rather than the source of transfer.

\subsection{Inductive-Bias-Oriented Pretraining}

Inductive-bias-oriented pretraining covers parametrically pretrained models whose transfer claims emphasize architecture or training design as much as corpus scale. TTM uses a compact mixer-based design for efficient zero-shot forecasting \cite{ekambaram2024ttm}; Mamba4Cast trains a state-space model solely on synthetic data for single-pass horizon prediction \cite{bhethanabhotla2024mamba4cast}; TiRex uses xLSTM state tracking and a masking strategy for in-context forecasting \cite{auer2025tirex}; and Reverso uses small hybrid convolution--linear-RNN models to improve the performance--efficiency trade-off \cite{fu2026reverso}. These mechanisms affect how transferable structure is stored and used, but the evidence still enters through parameters learned before downstream evaluation.

\subsection{Cross-Modal Alignment Pretraining}

Cross-modal alignment pretraining builds a shared interface between time-series tokens and language tokens. ChatTime models time series as a ``foreign language,'' expands the tokenizer, and uses continued pretraining together with instruction tuning to place numerical series and text in one model \cite{wang2025chattime}. It remains a parametric pretraining method because its transferable capability is learned into model weights before target evaluation, even if its language-interface features could also be disclosed.

\section{Retrieval-Augmented Memory}

Retrieval-augmented systems form a separate source class because external memory can change the evidence available at prediction time even when target-specific parameter updates are absent. This parallels the broader RAG distinction between parametric model memory and non-parametric retrieved memory \cite{lewis2020rag}, but forecasting requires additional disclosure about the retrieval store, retrieved objects, and fusion mechanism.

\subsection{Learnable Retrieval}

Learnable retrieval selects relevant external time-series examples for the current query while leaving the forecasting backbone fixed. TimeRAF combines a TSFM backbone, a learnable retriever, and Channel Prompting to integrate retrieved candidates from a task-specific time-series knowledge base \cite{zhang2025timeraf}. Although it uses a parametrically pretrained TSFM backbone, it belongs in the retrieval-augmented branch when retrieved candidates materially shape the reported forecast.

\subsection{Retrieved Pattern Fusion}

Retrieved pattern fusion treats retrieved examples as forecasting evidence rather than as passive context. TS-RAG retrieves semantically similar context--future pairs from a dedicated knowledge base and uses a learned augmentation module to fuse future-pattern evidence from retrieved pairs with the query representation \cite{ning2025tsrag}. The retrieval store therefore contributes both additional conditioning information and candidate future-pattern evidence.

\subsection{Query-Guided Retrieval Filtering}

Query-guided retrieval filtering makes relevance selection part of the forecasting computation. Cross-RAG uses query--retrieval cross-attention so that the query sequence can down-weight irrelevant retrieved examples and remain more stable as the retrieval budget grows \cite{lee2026crossrag}. Table~\ref{tab:retrieval-mechanisms} summarizes how the three retrieval designs differ in retrieval space, retrieved content, and fusion mechanism.

\begin{table}[t]
\caption{Retrieval evidence and fusion mechanisms.}
\label{tab:retrieval-mechanisms}
\scriptsize
\setlength{\tabcolsep}{3pt}
\renewcommand{\arraystretch}{0.9}
\resizebox{\columnwidth}{!}{%
\begin{tabular}{@{}llll@{}}
\toprule
\textbf{Method} & Retrieval space & Retrieved content used & Fusion mechanism \\
\midrule
\textbf{TimeRAF} & Learned embedding retrieval & Retrieved TS candidates & Channel Prompting \\
\textbf{TS-RAG} & Embedding: $e(q)\sim e(x)$ &
\makecell[l]{$y$ horizons from context\\--future pairs} &
Adaptive Retrieval Mixer \\
\textbf{Cross-RAG} & Data: $q\sim x$ & $(x,y)$ pairs & Query--retrieval cross-attention \\
\bottomrule
\end{tabular}%
}
\scriptsize\emph{Note.} $q$ denotes the query history, $x$ a retrieved historical window, $y$ its corresponding future horizon, and $e(\cdot)$ a retrieval encoder.
\end{table}

\section{Cross-Cutting Comparison}
\label{sec:comparison}

The taxonomy identifies the primary evidence source behind a zero-shot claim. It does not by itself fix the evaluation condition. Four audit questions remain: task interface, forecast object and scoring, prediction-time context, and resource budget. GIFT-Eval reports benchmark-side dimensions such as domain, frequency, variate structure, and prediction length \cite{aksu2024gifteval}; model-side comparisons need analogous disclosure. Table~\ref{tab:worked-audit-examples} applies the audit questions to the reported configurations of LLMTime, Chronos, and TimeRAF.

\begin{table}[t]
    \caption{Configuration-level applications of evidence-source and audit-condition disclosure.}
  \label{tab:worked-audit-examples}
  \centering
  \scriptsize
  \setlength{\tabcolsep}{3pt}
  \renewcommand{\arraystretch}{1.0}
  {
  \begin{tabularx}{\columnwidth}{@{}>{\raggedright\arraybackslash\bfseries}p{0.18\columnwidth}X@{}}
    \toprule
     Method & Evidence source and audit disclosures \\
    \midrule
    LLMTime & Evidence source: frozen LLM prior; interface: serialized numeric tokens; forecast object: samples or point reduction; context: target history only; resource issue: autoregressive decoding budget. \\
    \addlinespace[2pt]
    Chronos & Evidence source: parametric TS pretraining; interface: scaled and quantized tokens; forecast object: sampled forecast distribution; context: target history; resource issue: sampling budget. \\
    \addlinespace[2pt]
    TimeRAF & Evidence source: retrieval memory with a TSFM backbone; interface: TSFM input with retrieved candidates; forecast object: point or probabilistic output, depending on setup; context: target history plus retrieved series; resource issue: retrieval-store size and candidate count. \\
    \bottomrule
  \end{tabularx}}
\end{table}
\FloatBarrier

{\noindent\textbf{Task Interface.} How is the forecasting task represented to the model?}
A numerical history may be serialized as digit tokens, wrapped in forecasting-aware prompts, or scaled and quantized into a fixed vocabulary. In LLMTime, tokenization, scaling, precision, and context length are coupled; LSTPrompt changes the prompt construction; Chronos scales and quantizes values before autoregressive sampling \cite{gruver2024llmtime,liu2024lstprompt,ansari2024chronos}. {These choices define the task presented to the model; a score gain attributable to a different serialization scheme can be misread as a stronger transferable prior if the interface is not disclosed.}

{\noindent\textbf{Forecast Object and Scoring.} What predictive object is produced and scored?}
Point predictions and probabilistic forecasts represented by samples or quantiles support different claims. Common point-error measures can be scale-dependent or unstable in some settings \cite{hyndman2006accuracy}, while probabilistic forecasts should be evaluated with scoring rules appropriate to the predictive distribution \cite{gneiting2007proper}. {The scored object, any reduction applied to it, and the scoring rule therefore define the comparison; collapsing probabilistic samples to a median point estimate before scoring changes the predictive claim without changing the model, making metric ranks across forecast objects uninterpretable.}

{\noindent\textbf{Prediction-Time Context.} What information is available at prediction time?}
This axis records the information supplied for a particular forecast, not where the model acquired its general capability. MOIRAI supports arbitrary-variate inputs, ChatTime accepts numerical--text inputs, and TimeRAF supplies retrieved series at prediction time \cite{woo2024moirai,wang2025chattime,zhang2025timeraf}. {Methods within the same source class may therefore operate under different prediction-time information sets; a retrieval-augmented system and a parametrically pretrained model can share a leaderboard row while the former has access to matched external series the latter does not, inflating the apparent transfer gap.}

{\noindent\textbf{Resource Budget.} What resource budget is used to produce the forecast?}
Resource budgets also define comparisons: autoregressive sampling or decoding, sparse expert routing, compact inference, long-context state tracking, and retrieval or fusion costs place computation in different pipeline stages \cite{gruver2024llmtime,shi2025timemoe,ekambaram2024ttm,auer2025tirex,zhang2025timeraf}. {Accuracy-only rankings do not establish resource parity; a compact architecture reporting lower latency and a large autoregressive model reporting higher accuracy are not comparable without a shared resource account.}

\begin{table}[t]
  \caption{Minimum disclosures for interpreting zero-shot TSF scores.}
  \label{tab:min-disclosure}
  \centering
  \scriptsize
  \setlength{\tabcolsep}{3pt}
  \renewcommand{\arraystretch}{1.08}

  \begin{tabularx}{\columnwidth}{
    @{}
    >{\raggedright\arraybackslash\bfseries}p{0.28\columnwidth}
    >{\raggedright\arraybackslash}X
    @{}
  }
    \toprule
    Audit condition & Minimum disclosure \\
    \midrule
    Evidence access & Primary evidence source; material secondary sources; pretraining or retrieval provenance \\
    \addlinespace[2.5pt]
    Task interface & Encoding or prompt; normalization; usable history and horizon; generation mode \\
    \addlinespace[2.5pt]
    Forecast object \& scoring & Scored object; any point or distributional reduction; metric or scoring rule \\
    \addlinespace[2.5pt]
    Prediction-time context & Target history; available variates, covariates, text, or retrieved series \\
    \addlinespace[2.5pt]
    Resource budget & Model size; active parameters or experts; sampling or decoding budget; retrieval-store size and candidate count \\
    \bottomrule
  \end{tabularx}
\end{table}

\section{Evaluation Fragility and Research Agenda}
\label{sec:evaluation}

Evaluation becomes fragile when a zero-shot score omits either its evidence source or its evaluation conditions. LLMTime, Chronos, and TimeRAF illustrate why the same no-update label can encode frozen-prior reuse, parametric time-series pretraining, and retrieval-augmented memory \cite{gruver2024llmtime,ansari2024chronos,zhang2025timeraf}. Recent critiques of LLM-based forecasting, TSFM readiness, leakage, and cross-domain foundation behavior further motivate this conditional reading \cite{tan2024useful,zhang2025ready,meyer2025rethinking,karaouli2025foundational}. 
Accordingly, a leaderboard row should be interpreted as a conditional statement: under a specified evidence boundary and evaluation condition, model \(M\) obtains score \(S\). Accuracy alone cannot establish whether a difference reflects transferable capability or a different interface, context, scoring target, or resource budget.

 The visionary agenda is to make those conditions visible, comparable, and testable. Benchmark reports should pair accuracy with evidence-aware disclosure, including pretraining and retrieval provenance, prediction-time context, scored forecast object, scoring rule, and resource budget. More importantly, benchmark governance should separate comparison tracks by evidence access: closed-evidence runs that allow only target history and a declared pretrained model; retrieval-declared runs that publish the retrieval corpus, indexing time, candidate count, and fusion mechanism; and open-evidence runs that permit richer covariates, text, or external context while reporting those advantages explicitly. Within each track, diagnostic ablations should remove or freeze evidence channels---pretrained weights, serialization or prompt interface, retrieved examples, and additional covariates---to identify which source actually explains a gain.

A forward-looking zero-shot TSF benchmark would therefore treat each leaderboard row as a compact evidence contract rather than as a single score. This follows a broader lesson from empirical NLP: held-out scores alone are insufficient when development choices and computation budgets differ \cite{dodge2019show}. Such a benchmark also adapts the logic of data statements, where explicit provenance documentation supports more precise generalization claims \cite{bender2018data}. For TSF, the same logic should cover pretraining corpora, synthetic generators, benchmark splits, retrieval indexes, and covariate or text channels.

\section{Conclusion}

Zero-shot TSF should be interpreted through both evidence access and evaluation conditions. The source-first taxonomy distinguishes frozen LLM reuse, parametric time-series pretraining, and retrieval-augmented memory. The four audit dimensions record how each score is produced. Reporting both turns leaderboard entries into auditable evidence claims rather than standalone ranks.

\clearpage
\bibliographystyle{ACM-Reference-Format}
\bibliography{references}

@inproceedings{gruver2024llmtime,
  title = {Large Language Models Are Zero-Shot Time Series Forecasters},
  author = {Gruver, Nate and Finzi, Marc and Qiu, Shikai and Wilson, Andrew Gordon},
  booktitle = {Advances in Neural Information Processing Systems},
  volume = {36},
  pages = {19622--19635},
  year = {2023}
}

@inproceedings{liu2024lstprompt,
  title = {{LSTPrompt}: Large Language Models as Zero-Shot Time Series Forecasters by Long-Short-Term Prompting},
  author = {Liu, Haoxin and Zhao, Zhiyuan and Wang, Jindong and Kamarthi, Harshavardhan and Prakash, B. Aditya},
  booktitle = {Findings of the Association for Computational Linguistics: ACL 2024},
  pages = {7832--7840},
  publisher = {Association for Computational Linguistics},
  year = {2024}
}

@inproceedings{rasul2024lagllama,
  title = {{Lag-Llama}: Towards Foundation Models for Probabilistic Time Series Forecasting},
  author = {Rasul, Kashif and Ashok, Arjun and Williams, Andrew Robert and Ghonia, Hena and Bhagwatkar, Rishika and Khorasani, Arian and Darvishi Bayazi, Mohammad Javad and Adamopoulos, George and Riachi, Roland and Hassen, Nadhir and Bilo{\v{s}}, Marin and Garg, Sahil and Schneider, Anderson and Chapados, Nicolas and Drouin, Alexandre and Zantedeschi, Valentina and Nevmyvaka, Yuriy and Rish, Irina},
  booktitle = {NeurIPS 2023 Workshop on Robustness of Few-shot and Zero-shot Learning in Large Foundation Models},
  year = {2023},
  note = {R0-FoMo Workshop}
}

@inproceedings{das2024timesfm,
  title = {A Decoder-Only Foundation Model for Time-Series Forecasting},
  author = {Das, Abhimanyu and Kong, Weihao and Sen, Rajat and Zhou, Yichen},
  booktitle = {Proceedings of the 41st International Conference on Machine Learning},
  series = {Proceedings of Machine Learning Research},
  volume = {235},
  pages = {10148--10167},
  publisher = {PMLR},
  year = {2024}
}

@article{ansari2024chronos,
  title = {{Chronos}: Learning the Language of Time Series},
  author = {Ansari, Abdul Fatir and Stella, Lorenzo and Turkmen, Caner and Zhang, Xiyuan and Mercado, Pedro and Shen, Huibin and Shchur, Oleksandr and Rangapuram, Syama Sundar and Arango, Sebastian Pineda and Kapoor, Shubham and Zschiegner, Jasper and Maddix, Danielle C. and Wang, Hao and Mahoney, Michael W. and Torkkola, Kari and Wilson, Andrew Gordon and Bohlke-Schneider, Michael and Wang, Yuyang},
  journal = {Transactions on Machine Learning Research},
  year = {2024}
}

@inproceedings{woo2024moirai,
  title = {Unified Training of Universal Time Series Forecasting Transformers},
  author = {Woo, Gerald and Liu, Chenghao and Kumar, Akshat and Xiong, Caiming and Savarese, Silvio and Sahoo, Doyen},
  booktitle = {Proceedings of the 41st International Conference on Machine Learning},
  series = {Proceedings of Machine Learning Research},
  volume = {235},
  pages = {53140--53164},
  publisher = {PMLR},
  year = {2024}
}

@inproceedings{shi2025timemoe,
  title = {{Time-MoE}: Billion-Scale Time Series Foundation Models with Mixture of Experts},
  author = {Shi, Xiaoming and Wang, Shiyu and Nie, Yuqi and Li, Dianqi and Ye, Zhou and Wen, Qingsong and Jin, Ming},
  booktitle = {International Conference on Learning Representations},
  year = {2025},
  note = {Spotlight}
}

@inproceedings{ekambaram2024ttm,
  title = {Tiny Time Mixers ({TTMs}): Fast Pre-trained Models for Enhanced Zero/Few-Shot Forecasting of Multivariate Time Series},
  author = {Ekambaram, Vijay and Jati, Arindam and Dayama, Pankaj and Mukherjee, Sumanta and Nguyen, Nam H. and Gifford, Wesley M. and Reddy, Chandra and Kalagnanam, Jayant},
  booktitle = {Advances in Neural Information Processing Systems},
  volume = {37},
  pages = {74147--74181},
  year = {2024}
}

@inproceedings{bhethanabhotla2024mamba4cast,
  title = {{Mamba4Cast}: Efficient Zero-Shot Time Series Forecasting with State Space Models},
  author = {Bhethanabhotla, Sathya Kamesh and Swelam, Omar and Siems, Julien and Salinas, David and Hutter, Frank},
  booktitle = {NeurIPS 2024 Workshop on Time Series in the Age of Large Models},
  year = {2024},
  note = {TSALM Workshop Spotlight}
}

@inproceedings{auer2025tirex,
  title = {{TiRex}: Zero-Shot Forecasting Across Long and Short Horizons with Enhanced In-Context Learning},
  author = {Auer, Andreas and Podest, Patrick and Klotz, Daniel and B{\"o}ck, Sebastian and Klambauer, G{\"u}nter and Hochreiter, Sepp},
  booktitle = {Advances in Neural Information Processing Systems},
  volume = {38},
  year = {2025}
}

@inproceedings{fu2026reverso,
  title = {{Reverso}: Efficient Time Series Foundation Models for Zero-Shot Forecasting},
  author = {Fu, Xinghong and Li, Yanhong and Papaioannou, Georgios and Kim, Yoon},
  booktitle = {2nd ICML Workshop on Foundation Models for Structured Data},
  year = {2026}
}

@inproceedings{wang2025chattime,
  title = {{ChatTime}: A Unified Multimodal Time Series Foundation Model Bridging Numerical and Textual Data},
  author = {Wang, Chengsen and Qi, Qi and Wang, Jingyu and Sun, Haifeng and Zhuang, Zirui and Wu, Jinming and Zhang, Lei and Liao, Jianxin},
  booktitle = {Proceedings of the AAAI Conference on Artificial Intelligence},
  volume = {39},
  number = {12},
  pages = {12694--12702},
  year = {2025}
}

@article{zhang2025timeraf,
  title = {{TimeRAF}: Retrieval-Augmented Foundation Model for Zero-Shot Time Series Forecasting},
  author = {Zhang, Huanyu and Xu, Chang and Zhang, Yi-Fan and Zhang, Zhang and Wang, Liang and Bian, Jiang},
  journal = {IEEE Transactions on Knowledge and Data Engineering},
  volume = {37},
  number = {9},
  pages = {5654--5665},
  year = {2025}
}

@inproceedings{ning2025tsrag,
  title = {{TS-RAG}: Retrieval-Augmented Generation based Time Series Foundation Models are Stronger Zero-Shot Forecaster},
  author = {Ning, Kanghui and Pan, Zijie and Liu, Yu and Jiang, Yushan and Zhang, James Yiming and Rasul, Kashif and Schneider, Anderson and Ma, Lintao and Nevmyvaka, Yuriy and Song, Dongjin},
  booktitle = {Advances in Neural Information Processing Systems},
  volume = {38},
  year = {2025},
  note = {Poster}
}

@misc{lee2026crossrag,
  title = {Not All Retrievals are Useful: Cross-Attention for Input-Aware {RAG} in Time Series Forecasting},
  author = {Lee, Seunghan and Lee, Jaehoon and Seo, Jun and Yoo, Sungdong and Kim, Minjae and Lim, Tae Yoon and Kang, Dongwan and Choi, Hwanil and Lee, SoonYoung and Ahn, Wonbin},
  year = {2026},
  note = {arXiv preprint; official venue not verified}
}

@inproceedings{lewis2020rag,
  title = {Retrieval-Augmented Generation for Knowledge-Intensive {NLP} Tasks},
  author = {Lewis, Patrick and Perez, Ethan and Piktus, Aleksandra and Petroni, Fabio and Karpukhin, Vladimir and Goyal, Naman and Kuttler, Heinrich and Lewis, Mike and Yih, Wen-tau and Rockt{\"a}schel, Tim and Riedel, Sebastian and Kiela, Douwe},
  booktitle = {Advances in Neural Information Processing Systems},
  volume = {33},
  pages = {9459--9474},
  publisher = {Curran Associates, Inc.},
  year = {2020}
}

@techreport{bommasani2021foundation,
  title = {On the Opportunities and Risks of Foundation Models},
  author = {Bommasani, Rishi and others},
  institution = {Stanford Center for Research on Foundation Models},
  year = {2021}
}

@inproceedings{tan2024useful,
  title = {Are Language Models Actually Useful for Time Series Forecasting?},
  author = {Tan, Mingtian and Merrill, Mike A. and Gupta, Vinayak and Althoff, Tim and Hartvigsen, Thomas},
  booktitle = {Advances in Neural Information Processing Systems},
  volume = {37},
  pages = {60162--60191},
  year = {2024},
  note = {Spotlight}
}

@inproceedings{aksu2024gifteval,
  title = {{GIFT}-Eval: A Benchmark for General Time Series Forecasting Model Evaluation},
  author = {Aksu, Taha and Woo, Gerald and Liu, Juncheng and Liu, Xu and Liu, Chenghao and Savarese, Silvio and Xiong, Caiming and Sahoo, Doyen},
  booktitle = {NeurIPS 2024 Workshop on Time Series in the Age of Large Models},
  year = {2024},
  note = {TSALM Workshop}
}

@misc{meyer2025rethinking,
  title = {Rethinking Evaluation in the Era of Time Series Foundation Models: {(Un)known} Information Leakage Challenges},
  author = {Meyer, Marcel and Kaltenpoth, Sascha and Zalipski, Kevin and M{\"u}ller, Oliver},
  year = {2025},
  note = {arXiv preprint; official venue not verified}
}

@inproceedings{zhang2025ready,
  title = {Are Time Series Foundation Models Ready for Zero-Shot Forecasting?},
  author = {Zhang, Yunkai and Zeng, Qi and Zhang, Yawen and Xu, Zhijie and Zheng, Ming and Gao, Chongyang and Jiang, Muyan and Zheng, Zeyu},
  booktitle = {1st ICML Workshop on Foundation Models for Structured Data},
  year = {2025}
}

@inproceedings{karaouli2025foundational,
  title = {How Foundational are Foundation Models for Time Series Forecasting?},
  author = {Karaouli, Nouha and Coquenet, Denis and Fromont, Elisa and Mermillod, Martial and Reyboz, Marina},
  booktitle = {NeurIPS 2025 Workshop on Recent Advances in Time Series Foundation Models},
  year = {2025},
  note = {BERT2S Workshop}
}

@article{hyndman2006accuracy,
  title = {Another Look at Measures of Forecast Accuracy},
  author = {Hyndman, Rob J. and Koehler, Anne B.},
  journal = {International Journal of Forecasting},
  volume = {22},
  number = {4},
  pages = {679--688},
  year = {2006}
}

@article{gneiting2007proper,
  title = {Strictly Proper Scoring Rules, Prediction, and Estimation},
  author = {Gneiting, Tilmann and Raftery, Adrian E.},
  journal = {Journal of the American Statistical Association},
  volume = {102},
  number = {477},
  pages = {359--378},
  year = {2007}
}

@inproceedings{liang2024foundation,
  title = {Foundation Models for Time Series Analysis: A Tutorial and Survey},
  author = {Liang, Yuxuan and Wen, Haomin and Nie, Yuqi and Jiang, Yushan and Jin, Ming and Song, Dongjin and Pan, Shirui and Wen, Qingsong},
  booktitle = {Proceedings of the 30th ACM SIGKDD Conference on Knowledge Discovery and Data Mining},
  pages = {6555--6565},
  publisher = {Association for Computing Machinery},
  year = {2024}
}

@article{jin2026large,
  title = {Large Models for Time Series and Spatio-Temporal Data: A Survey and Outlook},
  author = {Jin, Ming and Kong, Yaxuan and Liang, Yuxuan and Zhang, Chaoli and Xue, Siqiao and Wang, Xue and Zhang, James and Wang, Yi and Chen, Haifeng and Li, Xiaoli and Tseng, Vincent S. and Zheng, Yu and Chen, Lei and Xiong, Hui and Pan, Shirui and Wen, Qingsong},
  journal = {ACM Computing Surveys},
  year = {2026},
  note = {Online first}
}

@inproceedings{dodge2019show,
  title = {Show Your Work: Improved Reporting of Experimental Results},
  author = {Dodge, Jesse and Gururangan, Suchin and Card, Dallas and Schwartz, Roy and Smith, Noah A.},
  booktitle = {Proceedings of the 2019 Conference on Empirical Methods in Natural Language Processing and the 9th International Joint Conference on Natural Language Processing ({EMNLP-IJCNLP})},
  pages = {2185--2194},
  publisher = {Association for Computational Linguistics},
  address = {Hong Kong, China},
  year = {2019},
  doi = {10.18653/v1/D19-1224},
  url = {https://aclanthology.org/D19-1224/}
}

@article{bender2018data,
  title = {Data Statements for Natural Language Processing: Toward Mitigating System Bias and Enabling Better Science},
  author = {Bender, Emily M. and Friedman, Batya},
  journal = {Transactions of the Association for Computational Linguistics},
  volume = {6},
  pages = {587--604},
  publisher = {MIT Press},
  year = {2018},
  doi = {10.1162/tacl_a_00041},
  url = {https://aclanthology.org/Q18-1041/}
}

\end{document}